\documentclass{article}
\usepackage{ijcai26}

\usepackage{times}
\usepackage{soul}
\usepackage{url}
\usepackage[hidelinks]{hyperref}
\usepackage[utf8]{inputenc}
\usepackage[small]{caption}
\usepackage{graphicx}
\usepackage{amsmath}
\usepackage{amsthm}
\usepackage{booktabs}
\usepackage{algorithm}
\usepackage{algorithmic}
\usepackage[switch]{lineno}
\usepackage{amsmath, amssymb, amsfonts}

\newcommand{\best}[1]{\textbf{#1}} 
\newcommand{\subbest}[1]{\underline{#1}} 

\title{S\textsuperscript{2}-VLA: State-Space Guided Vision-Language-Action Models for Long-Horizon Manipulation}

\author{
    Zhipeng Xie$^1$,
    Zongyi Han$^1$,
    Xiangyi Wei$^1$,
    Shiliang Sun$^2$,
    Yang Li$^1$ {\normalfont and}
    Jing Zhao$^{1*}$ 
\affiliations
    $^1$School of Computer Science and Technology, East China Normal University, Shanghai 200062, China\\
    $^2$State Key Laboratory of Submarine Geoscience, School of Automation and Intelligent Sensing, Shanghai Jiao Tong University, Shanghai 200240, China\\
\emails
    \{51285901044, 51285901051, 52285901014\}@stu.ecnu.edu.cn,
    shiliangsun@gmail.com,
    \{yli, jzhao\}@cs.ecnu.edu.cn
}

\begin{document}

\maketitle


\begin{abstract}
\setlength{\emergencystretch}{.5em} 
\noindent 

Vision-Language-Action (VLA) models have demonstrated strong capabilities in robotic manipulation, but their performance degrades significantly in long-horizon tasks due to cumulative error propagation. This limitation largely arises from static feature fusion mechanisms that rely on fixed weights to combine visual, language, and action representations, preventing the model from adapting to different phases of task execution. To address this limitation, we propose S\textsuperscript{2}-VLA, a framework that introduces a State-Space Guided Adaptive Attention (SSGAA) mechanism. SSGAA maintains a belief state that tracks task progression and generates dynamic gating weights to adaptively fuse information from three complementary sources visual features for spatial perception, task intents for high-level task planning, and temporal action sequences for execution consistency. This adaptive fusion allows the model to shift its focus throughout task execution, aligning with the evolving requirements of different task stages. Despite its compact 2B parameter size,  S\textsuperscript{2}-VLA consistently outperforms larger 7B-scale models and achieves state-of-the-art performance on long-horizon manipulation benchmarks, including LIBERO and SimplerEnv. highlighting the importance of adaptive feature fusion for long-horizon robotic manipulation.

\end{abstract}

\section{Introduction}

In recent years, significant advances in vision-language models (VLMs)~\cite{bai_qwen3-vl_2025,Chen_2024_CVPR,karamcheti_prismatic_2024} have propelled the development of generalist robotic systems capable of perception, reasoning, and action under open-ended instructions. Within this paradigm, Vision-Language-Action (VLA)~\cite{zhang_pure_2025} models have emerged as a promising framework that bridges high-level understanding to low-level control, enabling instruction driven robotic manipulation. The core of VLA research lies in effectively aligning multimodal representations comprising visual observations and language instructions with the action space to generate precise and executable policies.
\begin{figure}[t]
\centering
\includegraphics[width=0.46\textwidth]{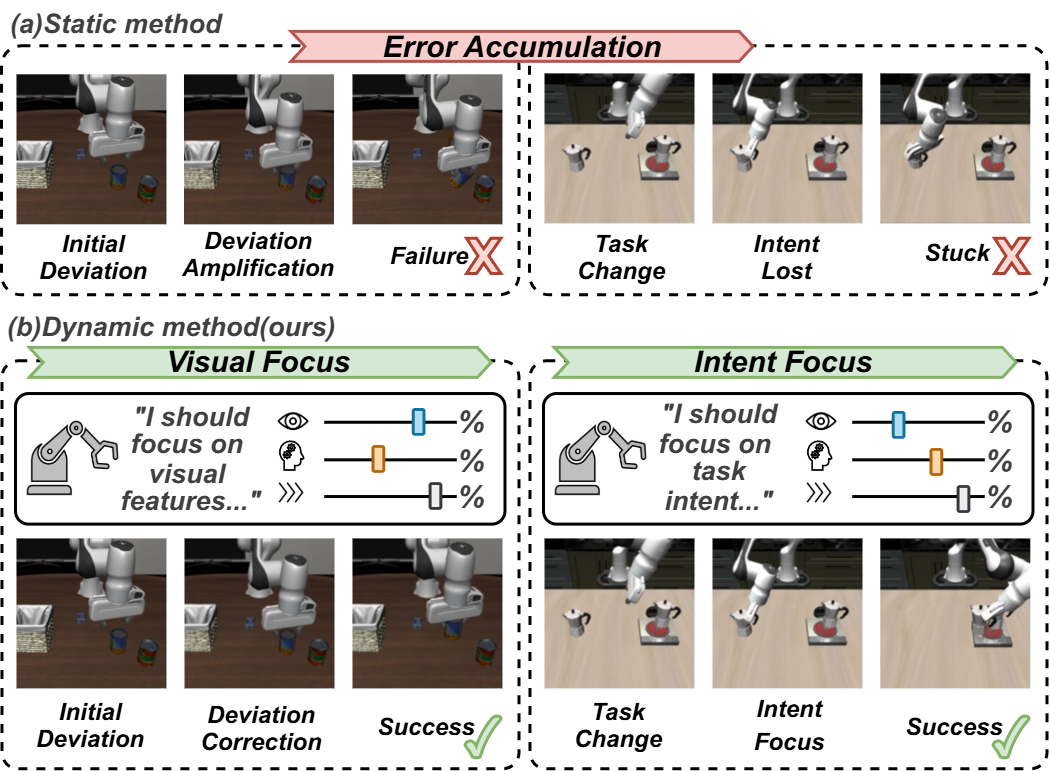}
\caption{
Illustrative examples comparing our S\textsuperscript{2}-VLA, which incorporates State-Space Guided Adaptive Attention, with static fusion approaches, highlighting its ability to mitigate early-stage bias.
}
\label{fig:ssg-aa-arch}
\end{figure}

The mainstream method aligns multimodal representations vision and language with the action space to generate executable policies by performing end-to-end fine-tuning of Vision-Language Models on large-scale robotic datasets~\cite{collaboration_open_2025,khazatsky_droid_2025}. However, when applied to long-horizon manipulation tasks that require multi-step reasoning, these models are fundamentally limited by their static representation fusion mechanism. Considering a task such as \textbf{put both the cream cheese box and the butter in the basket.} The model employs fixed weights to fuse information from different modalities like vision and language throughout its execution. Consequently, during the grasping phase which demands precise localization, the model cannot sufficiently focus on spatial details of the object. Meanwhile, in the planning phase which requires comprehension of the overall goal, it fails to allocate adequate attention to semantic intent. This lack of phase-adaptive capability in its rigid design leads to decision biases at critical steps. More critically, in long-horizon tasks, the lack of a context-aware correction mechanism allows these early biases to propagate and amplify along the action chain, ultimately leading to task failure.

To overcome this limitation, we propose the S\textsuperscript{2}-VLA framework. Its core is a novel State-Space Guided Adaptive Attention (SSGAA) mechanism. As illustrated in Figure~\ref{fig:ssg-aa-arch}, unlike static fusion, S\textsuperscript{2}-VLA maintains a compact belief state that tracks task progression and dynamically gates three complementary information streams spatial visual features, semantic task intent, and temporal action history enabling phase-aware adaptation. Extensive experiments on benchmarks such as LIBERO~\cite{liu2023libero}, SimplerEnv~\cite{li24simpler} and ALOHA~\cite{zhao_learning_2023} platform demonstrate that S\textsuperscript{2}-VLA, despite with compact 2B parameters, consistently outperforms larger 7B-scale models and sets a new state-of-the-art, effectively mitigating error propagation and significantly improving long-horizon task success.

The main contributions of our work are as follows:
 \begin{itemize}
    \item We propose S\textsuperscript{2}-VLA, which leverages pretrained VLMs to directly generate robot action policies through a belief-state-space interface, eliminating the need for large-scale action-sequence pretraining.
    \item We propose a novel State-Space Guided Adaptive Attention mechanism, which models task progress as a dynamic belief state. This belief state is used to guide a gating network that dynamically modulates the fusion of visual, linguistic, and action representations for long-horizon manipulation tasks.
    \item We show that S\textsuperscript{2}-VLA achieves robust long-horizon robotic manipulation with a lightweight model architecture, outperforming prior VLA methods on multiple standard benchmarks.
\end{itemize}

\section{Related Work}

\subsection{Vision-Language-Action Models}

In the ongoing development of generalist robotic policies, the field is observing a distinct trajectory that moves from imitation-learning-based expert policies toward general-purpose foundation models. Early pioneers in this domain, such as ACT~\cite{zhao_learning_2023}, proved that it is possible to achieve strong dexterity by directly fitting the action distribution from demonstration data. Following this breakthrough, researchers have further explored generalist policies pre-trained on large-scale robot trajectories, resulting in representative systems like Octo~\cite{team_octo_2024}, RDT~\cite{liu_rdt-1b_2025}, Seer~\cite{tian_predictive_2024}, RT-1~\cite{brohan_rt-1_2023}, RT-1-X~\cite{collaboration_open_2025}, and UnifiedVLA~\cite{li_unified_2025}. However, it is worth noting that these approaches often struggle to handle open-world semantic instructions effectively. They tend to face challenges with long-tail concepts and compositional generalization, especially in scenarios where semantic understanding is not grounded in strong pre-trained VLMs. Consequently, to address this bottleneck, contemporary research has started to leverage pre-trained VLMs as semantic backbones, a strategy that has successfully given rise to VLA models~\cite{zhang_pure_2025,gao_vla-os_2025}.

Current VLA architectures can be broadly summarized into two paradigms for action decoding. \emph{Autoregressive Token Generation Models} discretize actions into tokens and generate them sequentially, such as OpenVLA~\cite{kim_openvla_2024}, RT-2~\cite{zitkovich_rt-2_2023}, $\pi_0$-FAST~\cite{pertsch_fast_2025}, CoT-VLA~\cite{Zhao_2025_CVPR}, FlowVLA~\cite{zhong_flowvla_2025}, UniVLA~\cite{bu_univla_2025}, TraceVLA~\cite{zheng_tracevla_2025}, WorldVLA~\cite{cen_worldvla_2025}, SpatialVLA~\cite{qu_spatialvla_2025}, Magma~\cite{Yang_2025_CVPR}, MolmoAct~\cite{lee_molmoact_2025},  etc. While semantically more unified, this serial generation typically incurs higher inference latency. In contrast, \emph{Parallel Decoding with Decoupled Policy Heads} decouples representation from execution to simultaneously improve speed and precision, such as OpenVLA-OFT~\cite{kim_fine-tuning_2025}, $\pi_0$~\cite{black__0_2024}, GR00T N1~\cite{nvidia_gr00t_2025}, SmolVLA~\cite{shukor_smolvla_2025}, CogACT~\cite{li_cogact_2024}, 4D-VLA~\cite{zhang_4d-vla_2025}, ThinkAct~\cite{huang_thinkact_2025}, CronusVLA~\cite{li_cronusvla_2025}, MemoryVLA~\cite{shi_memoryvla_2025}, RoboVLMs~\cite{liu2025towards}, and others. Under this paradigm, the VLM mainly provides semantic representations or high-level planning signals, while concrete action generation is handled by a decoupled policy head (e.g., diffusion or flow-matching models or ACT-style chunking). We adopt the latter paradigm to achieve efficient continuous control, with a focus on optimizing the feature fusion mechanism, aiming to fully leverage the perceptual capabilities of VLMs and appropriately bridge them with action generation for effective long-horizon task execution.
\subsection{Policy Generation and Feature Integration}

The performance of decoupled VLAs depends on \textit{what} information is utilized from the backbone and \textit{how} it is integrated. Existing methods typically exhibit statically configured feature integration, which limits their adaptability.

\paragraph{Static Focus on Compressed Representations.} Most mainstream methods~\cite{liu_rdt-1b_2025,kim_fine-tuning_2025} restrict their attention to the final output representations. Specifically, OpenVLA-OFT~\cite{kim_fine-tuning_2025}  utilizes Learnable Action Queries to perform parallel decoding of future action chunks. However, the policy head attends exclusively to embeddings at these specific query token positions. While these embeddings encapsulate rich semantic intent, they constitute an information bottleneck, abstracting away the fine-grained spatial details present in the VLM's image token embeddings. This rigid focus makes it difficult for the model to recover the precise geometric information required for high-dexterity tasks. 

\paragraph{Static Multi-level Fusion.} To retrieve lost details, SmolVLA~\cite{shukor_smolvla_2025} and VLA-Adapter~\cite{wang_vla-adapter_2025} introduce mechanisms to aggregate multi-level backbone features (e.g., intermediate hidden states). Although this expands the information scope, the fusion strategy remains \textit{static}. The models learn a fixed set of weights to balance these information sources. Consequently, they fail to adaptively shift their focus as applying the same fusion ratio regardless of whether the robot is in a visually critical phase (e.g., aligning for a grasp) or an intent-critical phase (e.g., executing motion primitives).

\begin{figure*}[t!]
    \centering
     \includegraphics[width=0.8\textwidth]{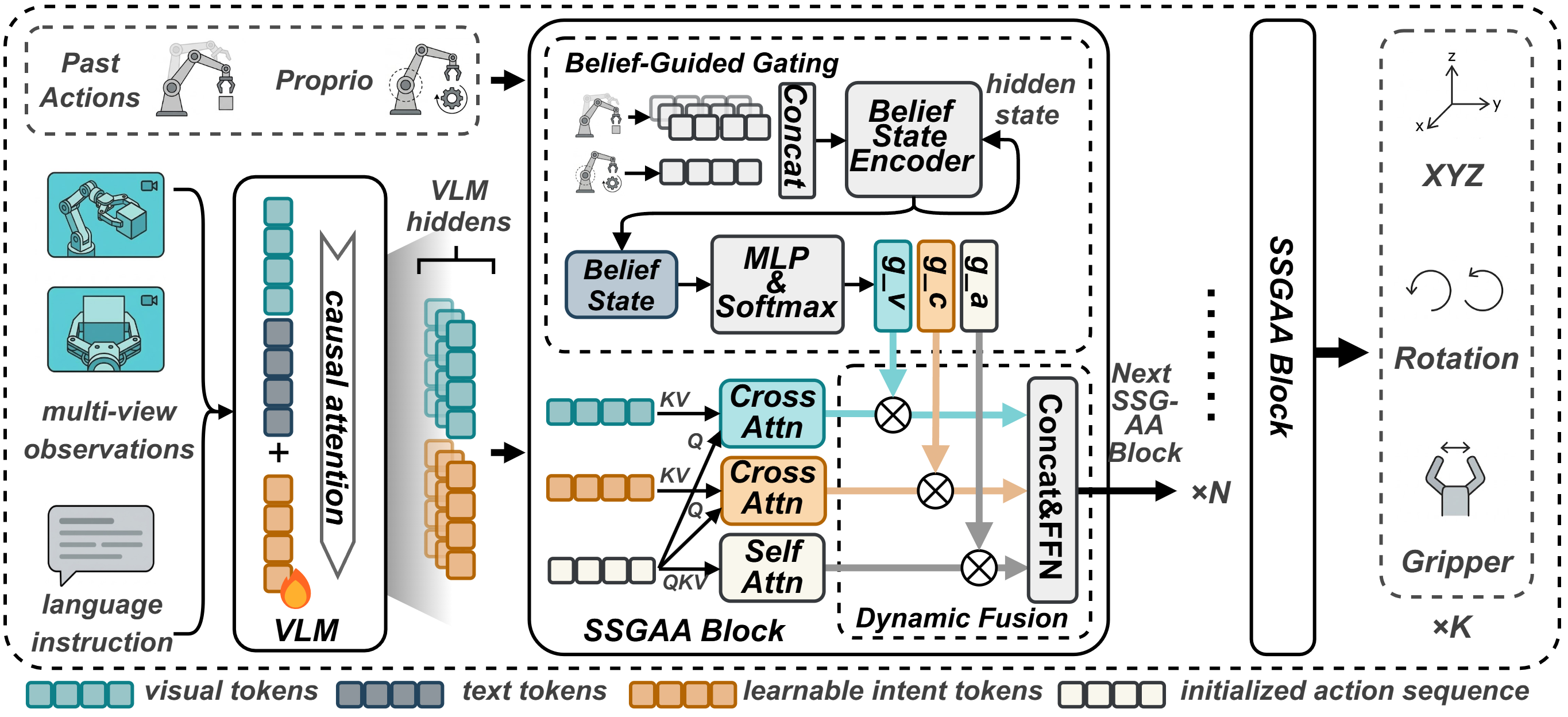} 
    \caption{Architecture and Workflow of the S\textsuperscript{2}-VLA Framework. The Workflow processes multimodal inputs through a vision-language backbone, with the core State-Space Guided Adaptive Attention(SSGAA) module adaptively fusing spatial, semantic, and temporal information under the dynamic guidance of the belief state. 
    }
    \label{fig:framework}
\end{figure*}

We argue that effective manipulation necessitates phase aware trade-offs visual precision for localization versus semantic intent for planning a critical temporal dynamic ignored by current methods.

\section{Methodology}
\label{sec:methodology}
\subsection{Problem Formulation}

We formalize the language-conditioned long-horizon manipulation tasks in VLA model. At each time  step \(t\), an agent receives a multi-view visual observation \(\mathbf{V}_t\), a natural language instruction \(\mathbf{L}_t\), and proprioceptive feedback \(\mathbf{P}_t\). A standard VLA policy is typically formulated as a direct mapping from these inputs to a future action sequence
\begin{equation}
    \mathbf{A}_{t:t+K-1} = \pi_{\text{VLA}}(\mathbf{V}_t, \mathbf{L}_t, \mathbf{P}_t),
    \label{eq:standard_vla}
\end{equation}
where \(\mathbf{A}_{t:t+K-1} = [\mathbf{a}_t, \mathbf{a}_{t+1}, \dots, \mathbf{a}_{t+K-1}]\) denotes \(K\) the predicted actions.

Although effective for short-horizon tasks, this formulation fundamentally lacks an explicit mechanism to maintain temporal coherence. Each action is generated from a static sensory snapshot, disregarding the execution history and how past actions have led to the current state. This makes the model susceptible to error accumulation in long-horizon scenarios.

\subsection{S\textsuperscript{2}-VLA}
\label{ssec:s2vla_framework}

\subsection*{Belief-Driven VLA Paradigm}

To overcome the lack of temporal coherence in static VLA models for long-horizon tasks, we formulate an internal belief state \(\mathbf{b}_t \in \mathbb{R}^{d_b}\). This state serves as a compact temporal context, summarizing \textbf{where we are} in task execution, and is recursively updated to encode task progression and historical information. The belief state update integrates the recent action sequence and its sensory feedback, formulated as
\begin{equation}
    \begin{cases}
        (\mathbf{o}_{t}^{(l)}, \mathbf{h}_{t}^{(l)}) = f_{\phi}(\mathbf{h}_{t}^{(l-1)}, \mathbf{A}_{t-K:t-1}, \mathbf{P}_t) \\
        \mathbf{b}_{t}^{(l)} = \mathbf{W}_b \cdot \mathbf{o}_{t}^{(l)} + \boldsymbol{\beta}_b
    \end{cases}
    \label{eq:belief_update}
\end{equation}
where \(f_{\phi}\) is a learned update function, \(K\) is the lookback horizon, and \(t\) and \(l\) denote the time step and the layer index of the action head, respectively. \(\mathbf{h}_{t}^{(l)}\) represents the hidden state updated via the \(l\) layer at time step \(t\), initialized by the low-level features \(\mathbf{h}_{t}^{(0)}\). This formulation explicitly models the causal relationship between the agent's actions \(\mathbf{A}_{t-K:t-1}\) and environmental feedback (proprioception), enabling the model to track task progress and detect execution deviations. For initial steps where the number of past actions is less than $K$, zero-padding is applied to the sequence. We implement \(f_{\phi}\) using a lightweight Gated Recurrent Unit (GRU)~\cite{cho_learning_2014}. Its recurrent nature maintains temporally coherent representations, compressing the history of action-perception pairs into a dynamic representation of task phase and execution quality. The complete policy is then conditioned on this belief state to generate actions as
\begin{equation}
    \mathbf{A}_{t:t+K-1} = \pi'_{\text{VLA}}(\mathbf{V}_t, \mathbf{L}_t, \mathbf{P}_t \; | \; \mathbf{b}_t).
    \label{eq:belief_conditioned_policy}
\end{equation}
The belief state \(\mathbf{b}_t\) is learned end-to-end from the action prediction loss, without any explicit supervision on task phases or execution errors. By minimizing prediction error, the model is forced to encode predictive information about future successful actions from the history of action-perception sequences. Consequently, distinct patterns arising from different task phases (e.g., \textbf{approaching}, \textbf{grasping}, \textbf{placing}) as well as discrepancies between expected and actual outcomes due to errors or perturbations are naturally encoded into \(\mathbf{b}_t\). This makes \(\mathbf{b}_t\) an latent representation that emergently encodes both task-phase information and execution fidelity, distilled directly from action-perception dynamics through end-to-end optimization.

\subsection*{Framework Overview}

The belief state \(\mathbf{b}_t\) provides crucial guidance for adaptive fusion of multimodal information. Building upon this, we propose the S\textsuperscript{2}-VLA framework. Its core is a novel State-Space Guided Adaptive Attention mechanism. SSGAA consist of three functionally complementary and parallel attention pathways. Instead of static fusion, these pathways are adaptively integrated via a dynamic gating network regulated by the belief state \(\mathbf{b}_t\), enabling stage-aware and adaptive feature aggregation. This design enables the policy to dynamically balance spatial details an d semantic intent based on action history and proprioception, thereby enhancing execution consistency in long-horizon tasks. The overall architecture of S\textsuperscript{2}-VLA are illustrated in Figure~\ref{fig:framework}.

\subsubsection*{Unified Input Construction}

Visual features $\mathbf{F}_t = \text{ViT}(\mathcal{V}_t) \in \mathbb{R}^{N_v \times d}$ are extracted from multi-view images $\mathcal{V}_t$. Language instructions $\mathbf{L}$ are tokenized into $\{\mathbf{l}_1, \mathbf{l}_2, \dots, \mathbf{l}_{N_l}\} \in \mathbb{R}^{N_l \times d}$, and learnable intent tokens $\mathbf{A}_{\text{tokens}}$ are embedded as $\{\tilde{\mathbf{a}}_1, \tilde{\mathbf{a}}_2, \dots, \tilde{\mathbf{a}}_{N_a}\} \in \mathbb{R}^{N_a \times d}$ for task-specific context. The combined input sequence is:
\begin{equation}
    \mathbf{X} = \bigl[ \mathbf{F}_t;\; \mathbf{l}_1, \dots, \mathbf{l}_{N_l};\; \tilde{\mathbf{a}}_1, \dots, \tilde{\mathbf{a}}_{N_a} \bigr] \in \mathbb{R}^{(N_v + N_l + N_a) \times d}.
    \label{eq:input_sequence}
\end{equation}
The sequence is processed through $H$ Transformer layers to produce hierarchical representations.
\subsubsection*{State-Space Guided Adaptive Attention}

The SSGAA mechanism consists of belief-guided dynamic gating and three complementary attention pathways, all operating on contextualized representations derived from the unified input sequence \(\mathbf{X}\).

\paragraph{Action Sequence Self-Attention.} Ensures temporal coherence in the predicted motion trajectories. This branch operates on a dedicated action-sequence embedding $\mathbf{A}^{(0)} \in \mathbb{R}^{T \times K \times d}$ where $T$ denotes the action dimension and $K$ the number of predicted steps, which is initialized as zero vectors and refined layer-wise through self-attention \cite{vaswani_attention_2017} within the SSGAA module. This enables modeling of the temporal dependencies across the predicted $K$ steps.

\paragraph{Low-Level Visual Cross-Attention.} Provides precise spatial grounding by attending to fine-grained visual features. We construct a visual vector \(\mathbf{C}_{\text{vis}} \in \mathbb{R}^{ N_v \times d}\) by stacking the hidden states of visual tokens from one transformer layer. The operation is defined as
\begin{equation}
    \mathbf{O}_{\text{vis}} = \operatorname{Softmax}\left(\frac{\mathbf{Q} (\mathbf{C}_{\text{vis}} \mathbf{W}_k^{\text{vis}})^{\!\top}}{\sqrt{d}}\right) (\mathbf{C}_{\text{vis}} \mathbf{W}_v^{\text{vis}}),
    \label{eq:visual_attention}
\end{equation}
where  \(\mathbf{Q}_{\text{}} \in \mathbb{R}^{ K \times d}\)is the query projection derived from the learnable action sequence representation and \(\mathbf{W}_k^{\text{vis}}, \mathbf{W}_v^{\text{vis}} \in \mathbb{R}^{d \times d}\) are learnable key and value projections that preserve fine-grained spatial details crucial for object manipulation.

\paragraph{High-Level Intent Cross-Attention.} Facilitates semantic intent understanding by processing aggregated multimodal context. We form a semantic context vector \(\mathbf{C}_{\text{ite}} \in \mathbb{R}^{N_a \times d}\) from the hidden states of the \(N_a\) learnable intent tokens across layers. The attention is computed as
\begin{equation}
    \mathbf{O}_{\text{ite}} = \operatorname{Softmax}\left(\frac{\mathbf{Q} (\mathbf{C}_{\text{ite}} \mathbf{W}_k^{\text{ite}})^{\!\top}}{\sqrt{d}}\right) (\mathbf{C}_{\text{ite}} \mathbf{W}_v^{\text{ite}}),
    \label{eq:semantic_attention}
\end{equation}
where \(\mathbf{W}_k^{\text{ite}}, \mathbf{W}_v^{\text{ite}} \in \mathbb{R}^{d \times d}\) project the action-token representations, which encode the distilled task semantics, into keys and values respectively.

\paragraph{Belief-Guided Gating.} The belief state \(\mathbf{b}_t\) dynamically modulates these three branches through a gating mechanism. At layer \(l\), soft gating weights are computed as
\begin{equation}
    \bigl[g_{\text{vis}}^{(l)},\; g_{\text{ite}}^{(l)},\; g_{\text{act}}^{(l)}\bigr]^{\!\top} = \operatorname{Softmax}\Bigl(\operatorname{MLP}_g^{(l)}(\mathbf{b}_{t}^{(l)}\Bigr).
    \label{eq:gating_weights}
\end{equation}
Branch outputs are fused as
\begin{equation}
    \mathbf{H}^{(l)} = g_{\text{vis}}^{(l)} \cdot \mathbf{O}_{\text{vis}}^{(l)} + g_{\text{ite}}^{(l)} \cdot \mathbf{O}_{\text{ite}}^{(l)} + g_{\text{act}}^{(l)} \cdot \mathbf{O}_{\text{act}}^{(l)},
    \label{eq:branch_fusion}
\end{equation}
where \(\mathbf{O}_{\text{vis}}^{(l)}\), \(\mathbf{O}_{\text{ite}}^{(l)}\), and \(\mathbf{O}_{\text{act}}^{(l)}\) denote the outputs of the visual, semantic, and action self-attention branches at layer \(l\), respectively. The fused representation \(\mathbf{H}^{(l)}\) is processed by a feed-forward network. The processed output then serves a dual function. First, it provides the query input \(\mathbf{Q}^{(l+1)}\) for the next layer's visual and context cross-attention branches. Second, it is fed into the same layer's action sequence self-attention branch. This design enables iterative refinement through multi-layer processing.

\subsubsection*{Parallel Action Decoding}
The final action prediction is generated by applying layer normalization followed by a linear projection to the output of the last SSGAA layer are computed as
\begin{equation}
    \hat{\mathbf{A}}_{t:t+K-1} = \operatorname{LN}\bigl(\mathbf{H}^{(L_{\text{out}})}\bigr)\mathbf{W}_{\text{out}}^\top + \boldsymbol{\beta}_{\text{out}},
    \label{eq:parallel_decoder}
\end{equation}
where \(\operatorname{LN}(\cdot)\) denotes layer normalization. The learnable parameters are the weight matrix \(\mathbf{W}_{\text{out}} \in \mathbb{R}^{T \times d}\) and the bias vector \(\boldsymbol{\beta}_{\text{out}} \in \mathbb{R}^{T}\). The bias \(\boldsymbol{\beta}_{\text{out}}\) is broadcasted across the temporal dimension \(K\) during the addition.

\subsection{Training and Inference Strategies}

\subsubsection*{Unified Loss Formulation}
The training objective is defined by the action prediction loss, which ensures precise trajectory generation by minimizing the discrepancy between predicted and ground-truth actions
\begin{equation}
    \mathcal{L}_{\text{total}} = \mathcal{L}_{\text{action}}\ =\frac{1}{K} \sum_{k=0}^{K-1} \|\mathbf{a}_{t+k} - \hat{\mathbf{a}}_{t+k}\|_2^2,
    \label{eq:action_loss}
\end{equation}
where $\mathbf{a}_{t+k}$ denotes the ground-truth action at future step $t+k$ and $\hat{\mathbf{a}}_{t+k}$ is the corresponding prediction. This formulation directly optimizes the model for action prediction accuracy without additional regularization terms.

\subsubsection*{End-to-End Training and Inference}

The S\textsuperscript{2}-VLA model is trained end-to-end, jointly optimizing all components including the pre-trained Qwen3-VL~\cite{bai_qwen3-vl_2025} backbone, belief update module, and SSGAA attention mechanism to allow gradient flow and coordinated adaptation for robotic manipulation. During inference, the model operates in a temporally recurrent loop. At each step, it first updates its state through a GRU~\cite{cho_learning_2014}. The model then computes adaptive gating weights to fuse visual, textual, and action features into a unified representation. Based on this representation, it predicts and executes the next action before proceeding to the following time step. This design ensures temporal consistency and robustness via continuous state estimation. Simultaneously, the dynamic gating mechanism enables automatic attention shifting across different modalities according to the current task phase, thereby facilitating adaptive and reliable task execution. Details are provided in the supplementary material.

\section{Experiments}

\subsection{Experimental Setup}
To evaluate the robustness of policies in long-horizon manipulation tasks, we evaluate S\textsuperscript{2}-VLA on three main benchmarks. During training, the model is optimized using four H100 GPUs. For inference, the system requires only 7GB of VRAM to perform deployment. In simulation, we use the LIBERO~\cite{liu2023libero} benchmark (Spatial, Object, Goal, and Long subsets) with fixed trajectories per task, language instructions, and third-person views, following official protocols with scripted initial scene sampling. Additionally, we evaluate within the SIMPLER~\cite{li24simpler} beachmark which is a high-fidelity simulation platform designed to bridge the real-to-sim control and visual gap by faithfully replicating real-world conditions for representative robotic arms such as WidowX. Extensive testing has demonstrated a strong correlation between SIMPLER performance and real-world results. In the real world, we use the ALOHA bimanual system~\cite{zhao_learning_2023} (30 Hz control) with multi-view overhead and wrist cameras and proprioceptive inputs. Tasks include pick-and-place, stacking, desktop organization, and utensil distribution. Experimental setup details are provided in the supplementary materials.
\begin{table}[t]
\centering
\resizebox{\linewidth}{!}{
\begin{tabular}{lcccccc}
\toprule
Method & \begin{tabular}[c]{@{}c@{}}Scale\\ (B)\end{tabular} & \begin{tabular}[c]{@{}c@{}}Spatial\\ (\%)\end{tabular} & \begin{tabular}[c]{@{}c@{}}Object\\ (\%)\end{tabular} & \begin{tabular}[c]{@{}c@{}}Goal\\ (\%)\end{tabular} & \begin{tabular}[c]{@{}c@{}}Long\\ (\%)\end{tabular} & \begin{tabular}[c]{@{}c@{}}Avg.\\ (\%)\end{tabular} \\
\midrule
FlowVLA (2025, ArXiv)         & 8.5  & 93.2 & 95.0  & 91.6 & 72.6 & 88.1  \\
UnifiedVLA (2025, ArXiv)       & 8.5  & 95.4 & 98.8 & 93.6 & 94.0 & 95.5  \\
OpenVLA (2024, CoRL)          & 7    & 84.7 & 88.4  & 79.2 & 53.7 & 76.5  \\
OpenVLA-OFT (2025, RSS) & 7    & \subbest{97.6}& 98.4  & \subbest{97.9} & \subbest{94.5}& \subbest{97.1}\\
UniVLA (2025, RSS)           & 7  & 96.5 & 96.8  & 95.6 & 92.0 & 95.2  \\
MemoryVLA (2025, ArXiv)        & 7  & \best{98.4} & 98.4  & 96.4 & 93.4 & 96.7  \\
CoT-VLA (2025, CVPR)          & 7    & 87.5 & 91.6  & 87.6 & 69.0 & 81.1  \\
WorldVLA (2025, ArXiv)         & 7    & 87.6 &\subbest{99.2} & 83.4 & 60.0 & 81.8  \\
CronusVLA (2025, AAAI)         & 7    & 97.3&\best{99.6} & 96.9 & 94.0 & 97.0  \\
TraceVLA (2025, ArXiv)         & 7    & 84.6 & 85.2  & 75.1 & 54.1 & 74.8  \\
MolmoAct (2025, ArXiv)         & 7    & 87.0 & 95.4  & 87.6 & 77.2 & 86.6  \\
ThinkAct (2025, NeurIPS)         & 7    & 88.3 & 91.4  & 87.1 & 70.9 & 84.4  \\
4D-VLA (2025, ArXiv)            & 4    & 88.9 & 95.2  & 90.9 & 79.1 & 88.6  \\
SpatialVLA (2025, RSS)        & 4    & 88.2 & 89.9  & 78.6 & 55.5 & 78.1  \\
$\pi_0$ (2025, RSS)           & 3    & 96.8 & 98.8 & 95.8 & 85.2 & 94.2  \\
$\pi_0$-FAST (2025, RSS)      & 3    & 96.4 & 96.8  & 88.6 & 60.2 & 85.5  \\
SmolVLA (2025, ArXiv)           & 2.2  & 93.0 & 94.0  & 91.0 & 77.0 & 88.8  \\
GR00T N1 (2025, ArXiv)          & 2    & 94.4 & 97.6  & 93.0 & 90.6 & 93.9  \\
Seer (2025, ArXiv)              & 0.57 & -    & -     & -    & 78.7 & 78.7  \\
VLA-OS (2025, ArXiv)            & 0.5  & 87.0 & 96.5  & 92.7 & 66.0 & 85.6  \\
\midrule
\textbf{S\textsuperscript{2}-VLA (ours)} & 2 & \best{98.4} & \best{99.6} & \best{98.4} & \best{96.4} & \best{98.2} \\
\bottomrule
\end{tabular}
}
\caption{Performance comparison on LIBERO benchmark (Spatial/Object/Goal/Long subsets).}
\label{tab:libero_comparison}
\end{table}

\subsection{Main Results Analysis}
\label{subsec:main_results}

\subsubsection{Performance on Simulation Benchmark}
\label{subsubsec:libero_results}
As shown in Table~\ref{tab:libero_comparison}, we conduct comparative experiments with multiple methods on the LIBERO benchmark. \best{Bold} is the best performance and \subbest{Bold} is the suboptimal performance. The results indicate that S\textsuperscript{2}-VLA demonstrates SOTA while maintaining efficiency advantages. It achieves success rates of 98.4\%, 99.6\%, 98.4\%, and 96.4\% on the Spatial, Object, Goal, and Long-Horizon subsets, respectively, with an average success rate of 98.2\%, significantly outperforming the compared methods. Particularly in long-horizon tasks, the model effectively mitigates error accumulation issues through the State-Space Guided Adaptive Attention mechanism.

Table~\ref{tab:memoryvla_calvin} presents the performance on four manipulation tasks using the WidowX robot platform. Our method still achieves state-of-the-art results even in the highly realistic simulation environment, SimplerEnv-Bridge.

\begin{table}[t]
\centering
\resizebox{\linewidth}{!}{
\begin{tabular}{lccccc}
\toprule
Method & \begin{tabular}[c]{@{}c@{}}Spoon\\ on Towel\end{tabular} & \begin{tabular}[c]{@{}c@{}}Carrot\\ on Plate\end{tabular} & \begin{tabular}[c]{@{}c@{}}Stack\\ Cube\end{tabular} & \begin{tabular}[c]{@{}c@{}}Eggplant\\ in Basket\end{tabular} & \begin{tabular}[c]{@{}c@{}}Avg.\\ Success\end{tabular} \\
\midrule
RT-1-X (2024, ICRA)       & 0.0  & 4.2  & 0.0  & 0.0   & 1.1  \\
OpenVLA (2024, RSS)       & 4.2  & 0.0  & 0.0  & 12.5  & 4.2  \\
Octo-Base (2024, ArXiv)   & 15.8 & 12.5 & 0.0  & 41.7  & 17.5 \\
RoboVLMs (2024, ArXiv)        & 45.8 & 20.8 & 4.2  & 79.2  & 37.5 \\
CogACT-Base (2024, ArXiv)       & 71.7 & 50.8 & 15.0 & 67.5  & 51.3 \\
CogACT-Large (2024, ArXiv)    & 58.3 & 45.8 & 29.2 & \subbest{95.8}  & 57.3 \\
$\pi_0$-Uniform* (2024, ArXiv) & 63.3 & 58.8 & 21.3 & 79.2  & 55.7 \\
$\pi_0$-Beta* (2024, ArXiv) & \best{84.6} & 55.8 & \best{47.9} & 85.4  & 68.4 \\
Magma (2025, PP)    & 37.5 & 29.2 & 20.8 & 91.7  & 44.8 \\
TraceVLA (2025, ArXiv)    & 12.5 & 16.6 & 16.6 & 65.0  & 27.7 \\
SpatialVLA (2025, ArXiv)      & 16.7 & 25.0 & 29.2 & \best{100.0} & 42.7 \\
MemoryVLA (2025, ArXiv)        & 75.0 & \subbest{75.0} & 37.5 & \best{100.0} & \textbf{71.9} \\
\midrule
\textbf{S\textsuperscript{2}-VLA (ours)} & \subbest{83.3} & \best{87.5} & \textbf{41.7} & \best{100.0} & \best{78.1} \\
\bottomrule
\end{tabular}
}
\caption{Performance comparison on SimplerEnv-Bridge with WidowX robot.}
\label{tab:memoryvla_calvin}
\end{table}

\subsubsection{Real-World Experimental Evaluation}

We evaluated S\textsuperscript{2}-VLA on a real-world ALOHA dual-arm mobile manipulation platform through a series of complex manipulation tasks. The evaluation comprised four distinct tasks. In the pick-and-place task, the positions of a blue and a yellow block were randomly initialized, requiring the system to pick up the blue block and place it into a bowl. In the stacking task, after random initialization of the block positions, the system was required to stack the yellow block on top of the blue block. The desktop organization task involved randomly initializing block positions and then first placing the yellow block into a bowl, followed by placing the blue block into the same bowl. Finally, the utensil distribution task required the system, after random initialization of utensil positions, to use the left arm to pick up a pair of chopsticks, pass them to the right arm, and then place the chopsticks into a bowl with the right arm. Details are provided in supplementary materials.

\begin{figure}[t]
\centering
\includegraphics[width=\linewidth]{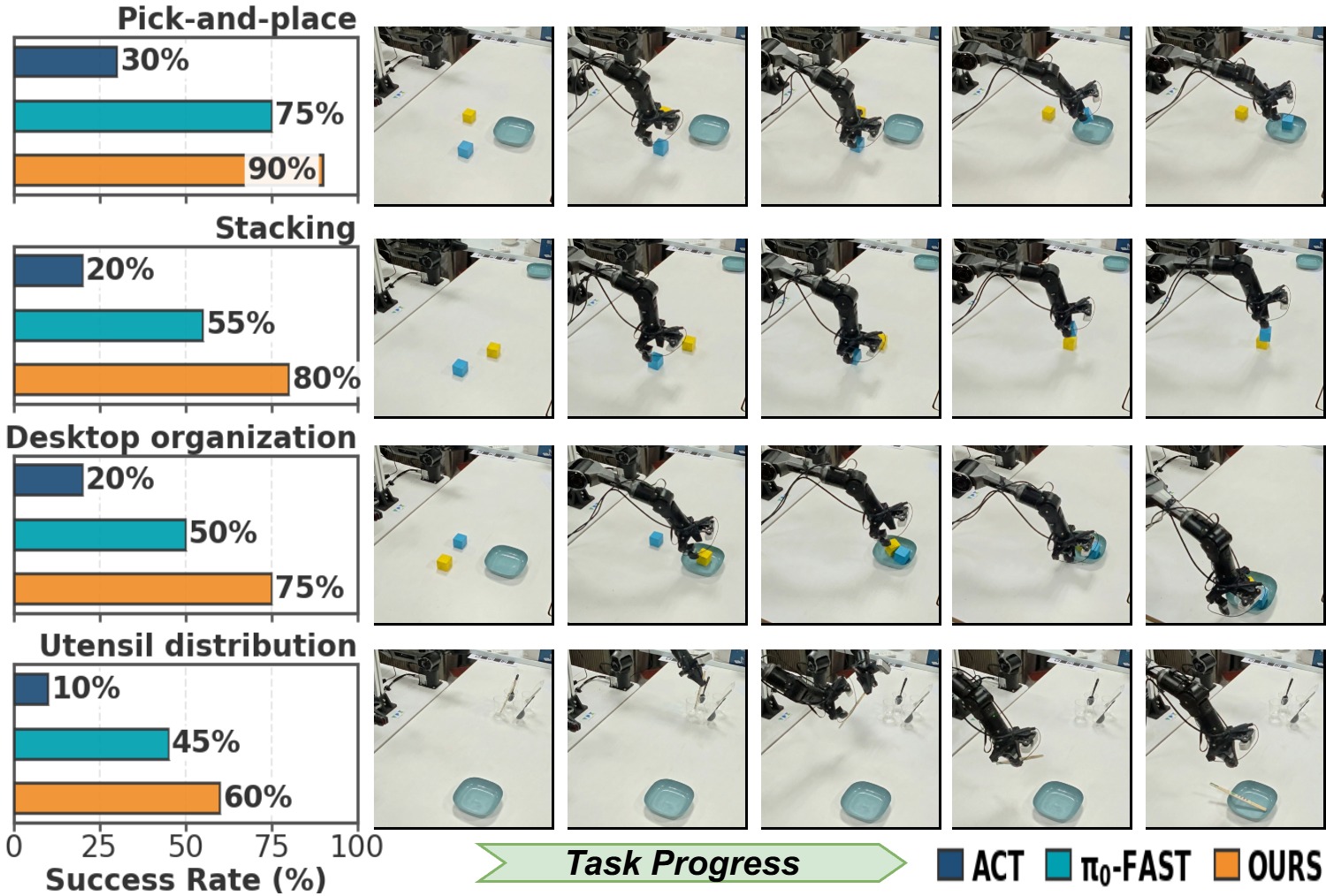}
\caption{Real World Performance of our S\textsuperscript{2}-VLA compared with ACT and $\pi_0$-FAST.}
\label{fig:real-world-perf}
\end{figure}
As illustrated in Figure~\ref{fig:real-world-perf}, the key steps in the execution process of our S\textsuperscript{2}-VLA are presented. It achieved superior success rates compared to all baseline methods, demonstrating that its SSGAA mechanism enables effective real-world transfer capability.
\subsection{Ablation Studies}
\label{subsec:ablation}

This section ablates the core design of S\textsuperscript{2}-VLA, the action head constructed from a 24-layer SSGAA mechanism. We systematically address four key questions \textit{(i) Is dynamic gating necessary? (ii) Is gating performance depth-sensitive? (iii) Does belief-state guidance provide additional benefit? (iv) What is the optimal SSGAA configuration? } All experiments follow the same training and evaluation protocols as the main experiments, with only the gating strategy modified.

\begin{table}[t]
\centering
\resizebox{\linewidth}{!}{
\begin{tabular}{lccc}
\toprule
Setting & Gated layers & Perf.(\%) & $\Delta$ vs.\ w/o Gate \\
\midrule
w/o Gate & -- & 95.0 & +0.0 \\
Gate@All & all layers & 94.4 & -0.6 \\
\midrule
\multicolumn{4}{l}{\textbf{Single-layer ablations}} \\
Gate@1  & 1  & 94.2 & -0.8 \\
Gate@6  & 6  & 96.0 & +1.0 \\
Gate@12 & 12 & 96.4 & +1.4 \\
Gate@18 & 18 & 95.8 & +0.8 \\
Gate@24 & 24 & 94.8 & -0.2 \\
\midrule
\multicolumn{4}{l}{\textbf{Combination strategy}} \\
Combination & 6,12,18 & 94.4 & -0.6 \\
\midrule
\multicolumn{4}{l}{\textbf{State guidance ablation}} \\
w/o State & 12 & 95.8 & +0.8 \\
\midrule
\textbf{S\textsuperscript{2}-VLA(ours)} & \textbf{12} & \textbf{96.4} & \textbf{+1.4} \\
\bottomrule
\end{tabular}
}
\caption{Ablation on gating configuration and belief-state guidance. $\Delta$ shows performance change relative to "w/o Gate".}
\label{tab:gate_layer_ablation}
\end{table}
The study begins by evaluating whether state-driven gating improves performance. As shown in Table~\ref{tab:gate_layer_ablation}, disabling gating (\textbf{w/o Gate}) results in 95.0\% performance, while enabling gating in all layers (\textbf{Gate@All}) slightly degrades performance to 94.4\%. This suggests that indiscriminate gating across all layers is not beneficial. To investigate depth sensitivity, we conduct single-layer gating experiments, enabling gating only at specific layers while keeping others static. Results reveal a clear pattern that gating provides significant gains at middle layers (6, 12, 18), with layer 12 achieving the best performance (96.4\%), while shallow and deep layers show minimal or negative impact. This reflects the hierarchical structure of feature extraction. Intermediate levels capture semantically rich features ideal for adaptive resource allocation, while early and final stages are better by static fusion.

Given the success of single-layer gating at layer 12, we investigate whether combining multiple high-performing layers (6, 12, 18) yields further improvements. Surprisingly, this multi-layer configuration performs worse (94.4\%) than both the best single-layer setting and the no-gating baseline. This indicates that multi-point gating introduces instability. To isolate the contribution of belief-state guidance, we replace the dynamic state input \(\mathbf{b}_t\) with a learnable static vector in the Gate@12 configuration. This \textbf{w/o State} variant achieves only 95.8\%, significantly lower than the full SSGAA (96.4\%). This confirms that performance gains arise not merely from introducing gating, but specifically from the dynamic modulation of gating weights by the evolving belief state, enabling phase-aware adaptation.
    
\subsubsection*{Ablation on State Guided Gating}
To further validate the core role of belief-state guidance, we conduct a controlled experiment. On the best-performing gating layer (Gate@12), we remove the input from the belief state \(\mathbf{b}_t\) and replace it with a learnable constant vector to generate static gating weights (denoted as \textbf{w/o Belief State}). The results in table~\ref{tab:gate_layer_ablation} show that the static gating setup performs worse than the full belief-state-guided gating. This demonstrates that the performance improvement is not simply due to the introduction of a gating mechanism, but rather depends critically on having the gating weights be dynamically modulated by the current belief state \(\mathbf{b}_t\). This dynamic mechanism enables the model to adaptively adjust the fusion of multi-source information (visual, semantic, and action-sequence) according to the task phase.
Our ablation study demonstrates that Gating is more effective at semantically rich middle layers and belief-state guidance is essential for optimal performance. Consequently, S\textsuperscript{2}-VLA adopts single-layer gating at a critical middle layer (layer 12, The experimentally determined optimal SSGAA configuration layer) with belief-state guidance as its core configuration.

\subsection{Evaluation of QwenVL-Based Models}
\label{subsec:qwenvl_sota}
\begin{table}[t]
\centering
\resizebox{\linewidth}{!}{
\begin{tabular}{lccccc}
\toprule
Method  / Model & Size & SR(\%) & $\Delta$ vs.\ $S^2$-VLA \\
\midrule
Qwen2.5-VL + FAST (2025, RSS)  & 3B & 90.2 & -6.2 \\
Qwen2.5-VL + OFT (2025, RSS)   & 3B & 92.0 & -4.4 \\
Qwen2.5-VL + PI (2025, RSS)    & 3B & 88.4 & -8.0 \\
Qwen2.5-VL + GR00T (2025, ArXiv) & 3B & 90.8 & -5.6 \\
\midrule
Qwen3-VL + FAST (2025, RSS)  & 4B & 90.6 & -5.8 \\
Qwen3-VL + OFT (2025, RSS)   & 4B & 93.8 & -2.6 \\
Qwen3-VL + PI (2025, RSS)    & 4B & 88.4 & -8.0 \\
Qwen3-VL + GR00T (2025, ArXiv) & 4B & 92.0 & -4.4 \\
\midrule
Qwen3-VL + Adapter-Pro (2025, AAAI) & 2B & 94.4 & -2.0 \\
\textbf{S\textsuperscript{2}-VLA (ours)} & \textbf{2B} & \textbf{96.4} & \textbf{--} \\
\bottomrule
\end{tabular}
}
\caption{Performance comparison on  LIBERO Long benchmark.}
\label{tab:qwenvl_sota}
\end{table}

\begin{figure*}[t!]
    \centering
    \includegraphics[width=1\textwidth]{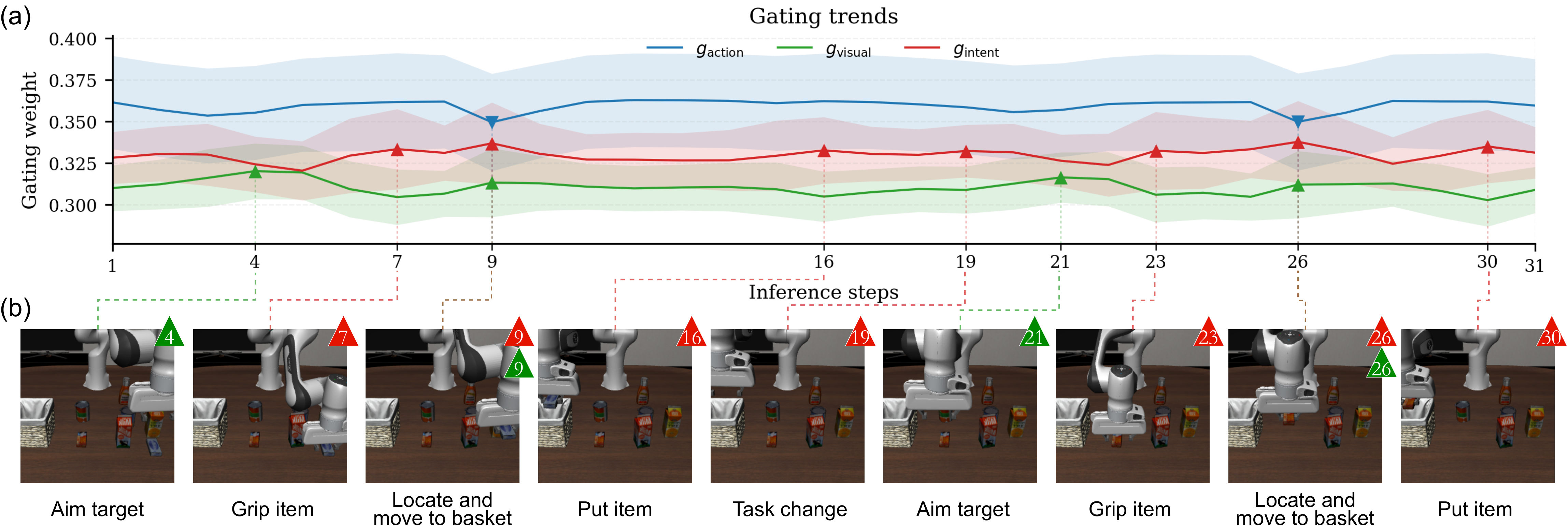}
    \caption{Gating weight trajectories and stage-aligned keyframes for a representative LIBERO-Long rollout.}
    \label{fig:gating_weights}
\end{figure*}

To isolate the performance gains attributable to architectural design, we systematically transplant several mainstream SOTA VLA Policy Head methods onto the QwenVL family of backbones including Qwen2.5-VL~\cite{bai_qwen25-vl_2025} and Qwen3-VL~\cite{bai_qwen3-vl_2025} for a fair comparison. As shown in Table~\ref{tab:qwenvl_sota}, although all methods are built upon the same pre-trained QwenVL weights, S\textsuperscript{2}-VLA achieves the highest success rate of 96.4\% with its compact 2B parameter architecture. This performance not only surpasses the similarly sized Qwen3-VL+Adapter-Pro~\cite{wang_vla-adapter_2025} with the success rate of 94.4\% but also exceeds all methods using larger 3B and 4B backbones, which Qwen3-VL+OFT~\cite{kim_fine-tuning_2025} only gets success rate of 93.8\%. The results demonstrate that the SSGAA mechanism is more effective than merely scaling up parameters for long-horizon tasks. S\textsuperscript{2}-VLA successfully breaks through the performance plateau observed in static fusion methods, which typically saturates in the 88\%-94\% range as task complexity increases. By integrating the semantic representations from the foundation model with real-time state predictions, it achieves dynamic shifting of perceptual focus across task phases, thereby enhancing robustness against long-horizon error accumulation.

\subsection{Gating Interpretability Analysis}

The primary cause of failure in long-horizon manipulation tasks lies in the propagation and amplification of perception or decision-making errors from early key steps during subsequent execution, ultimately leading to overall task failure. To elucidate how S\textsuperscript{2}-VLA addresses this challenge at a mechanistic level, we performed step-by-step visualization of the gating weights in its SSGAA module. A representative long-horizon task \textbf{put both the cream cheese box and the butter in the basket} is selected for analysis. During a successful execution trajectory, the model’s dynamic allocation of gating weights across its three attention branches is recorded at each reasoning step, which include low-level visual cross-attention (preserving fine-grained spatial details), high-level semantic cross-attention (goal-oriented planning), and action sequence self-attention (temporal coherence in motion generation). The resulting dynamic trajectories of the weights, aligned with stage-wise keyframes, are shown in Figure~\ref{fig:gating_weights}. The main observations are summarized as follows.

During phases that require precise object grounding and fine spatial alignment (e.g., when the end-effector approaches the target object, corresponding to the “Aim target” and “Locate and move to basket” keyframes marked by a green triangle in Figure~\ref{fig:gating_weights}), the weight of the low-level visual cross-attention (green curve, $g_\text{visual}$) rises significantly. This indicates that the model relies more heavily on early-layer visual tokens to support high-precision positioning.
    
At key decision points involving subtask switching or phase transitions (e.g., triggering the grasp action, transporting the object to the basket, and executing the release action, corresponding to stages such as “Grip item” and “Place item” in keyframes marked by a red triangle Figure~\ref{fig:gating_weights}), the weight of the high-level Intent cross-attention (red curve, $g_\text{intent}$) exhibits clear peaks. This demonstrates that the model leverages aggregated multimodal task representations to perform explicit stage transitions and behavioral planning.
    
During long-range, steady-state motion segments, the action sequence self-attention (blue curve, $g_\text{action}$) remains dominant. This allows the model to maintain smooth and temporally consistent trajectories through coherent step-by-step motion modeling, thereby reducing unnecessary perceptual and planning overhead.

This SSGAA mechanism enables the model to adaptively enhance the most pertinent information-processing pathway according to different task phases, thereby improving decision robustness in error-prone stages such as fine manipulation and phase switching. Experimental results confirm that this mechanism effectively reduces the likelihood of errors in critical steps, suppresses error propagation and accumulation from the source, and consequently enhances both the overall success rate and execution stability of long-horizon manipulation tasks.

\subsection{Performance Efficiency Metrics}
To verify inference efficiency, we compare the 7B model with our method, with results shown in Table~\ref{tab:efficiency_comparison}.

\begin{table}[t]
\centering
\resizebox{\linewidth}{!}{
\begin{tabular}{lcccc}
\toprule
\textbf{Efficiency} & TraceVLA & CronusVLA & OpenVLA-OFT & \textbf{S\textsuperscript{2}-VLA(ours)} \\
\midrule
Throughput (Hz) $\uparrow$ & 4.3 & 8.7 & \subbest{71.4} & \best{80.8} \\
\bottomrule
\end{tabular}
}
\caption{Inference efficiency comparison.}
\label{tab:efficiency_comparison}
\end{table}

\section{Conclusion}
S\textsuperscript{2}-VLA effectively mitigates error accumulation in long-horizon tasks by employing a state-space-guided adaptive attention mechanism, which dynamically fuses visual, linguistic, and action representations. Despite its compact size of 2B parameters, the model outperforms conventional 7B-scale models and achieves state-of-the-art success rates on long-horizon manipulation benchmarks, including  LIBERO~\cite{liu2023libero} and SimplerEnv~\cite{li24simpler}. Future work will explore the architectural compatibility of the SSGAA mechanism and extend its core idea of belief state modeling to different generative frameworks, such as Diffusion models or Flow Matching, to enhance the modeling capability for complex multimodal action distributions.

\section*{Acknowledgments}
This work is supported by the National Natural Science Foundation of China under Projects 62476089, 62576206 and 62572194, and by the Shanghai Science and Technology Committee under Grant No. 24511103202.

\section*{Contribution Statement}
Zhipeng Xie and Zongyi Han contributed equally to this work. Jing Zhao is the corresponding author.

\bibliographystyle{named}
\bibliography{ijcai26}

\end{document}


\maketitle

\appendix

\section{Setup Details of Simulation Benchmarks}

\begin{figure}[H]
\centering
\includegraphics[width=0.45\textwidth]{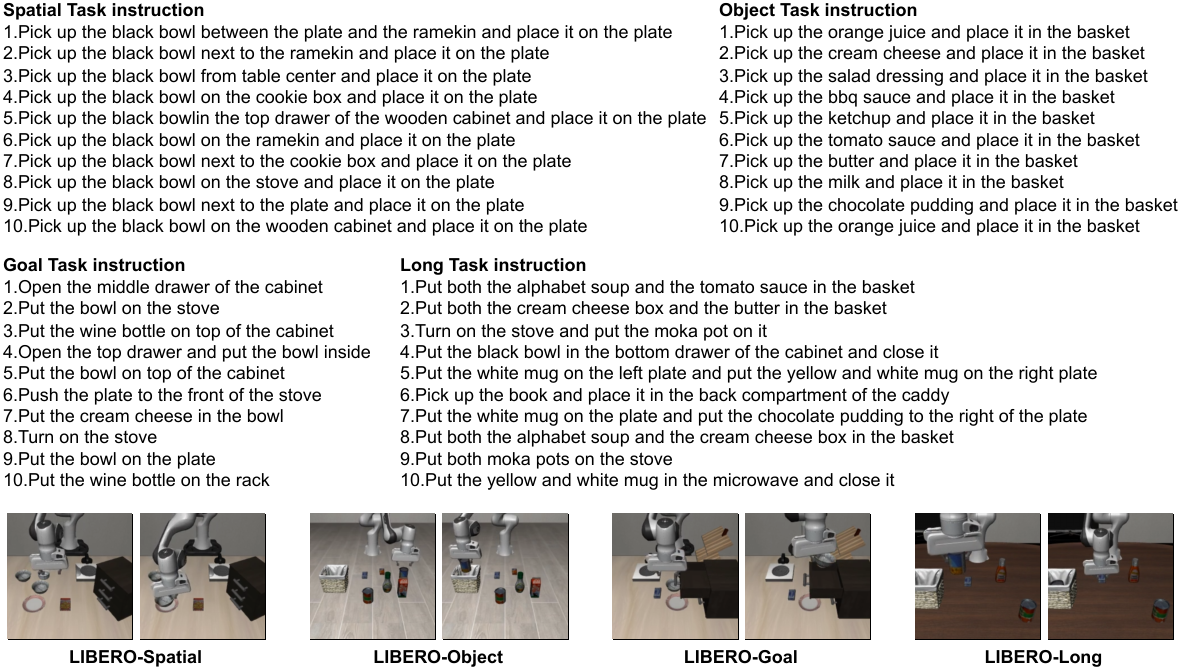}
\caption{Example tasks and natural language instructions from the LIBERO benchmark, which evaluates robotic manipulation across diverse task structures. Each visual observation is paired with its defining language instruction.}
\label{fig:libero}
\end{figure}

The LIBERO benchmark~\cite{liu2023libero} is designed to evaluate robotic manipulation policies across diverse task structures. The LIBERO benchmark comprises four distinct task suites, including LIBERO-Spatial, LIBERO-Object, LIBERO-Goal, and LIBERO-100. The first three suites each contain 10 tasks that focus on specific reasoning challenges, such as spatial relationships, object affordances, and procedural goals. LIBERO-100 is a larger suite containing 90 short-horizon tasks (LIBERO-90) and 10 complex long-horizon tasks (LIBERO-Long), which are particularly crucial for assessing error propagation in multi-step reasoning. In this benchmark, the strategy for each task is generated open-loop, depending solely on the language instruction provided at the onset, without additional environmental feedback. To ensure statistical reliability, each task is executed for 50 repetitions with varied initial conditions, and the average success rate is reported for each subtask. Example scenarios and their corresponding natural language instructions from the LIBERO benchmark are illustrated in Figure~\ref{fig:libero}.


\begin{figure}[t]
\centering
\includegraphics[width=0.45\textwidth]{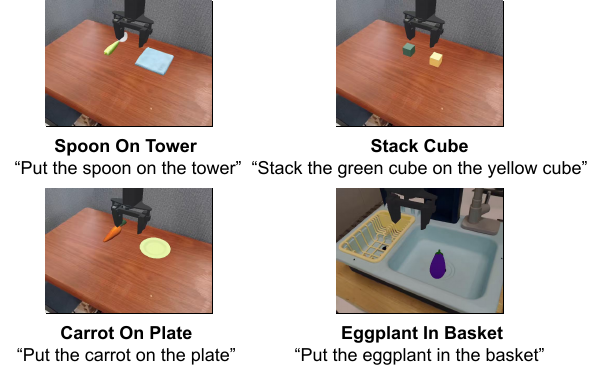}
\caption{Task execution in the SimplerEnv-Bridge environment. The tasks include Spoon on Towel, Carrot on Plate, Stack Cube, and Eggplant in Basket. All experiments are conducted on the WidowX robotic platform.}
\label{fig:simpler}
\end{figure}

To assess the robustness of the model, we conduct additional experiments in the SimplerEnv-Bridge simulation environment~\cite{li24simpler}. The evaluation includes the tasks Spoon on Towel, Carrot on Plate, Stack Cube, and Eggplant in Basket. For statistical reliability, each task is executed 24 repetitions under varied initial conditions, and the average success rate of each subtask is recorded. All tasks are performed on the WidowX robot platform, and the corresponding experimental setups are shown in Figure~\ref{fig:simpler}.

\begin{figure}[H]
\centering
\includegraphics[width=0.45\textwidth]{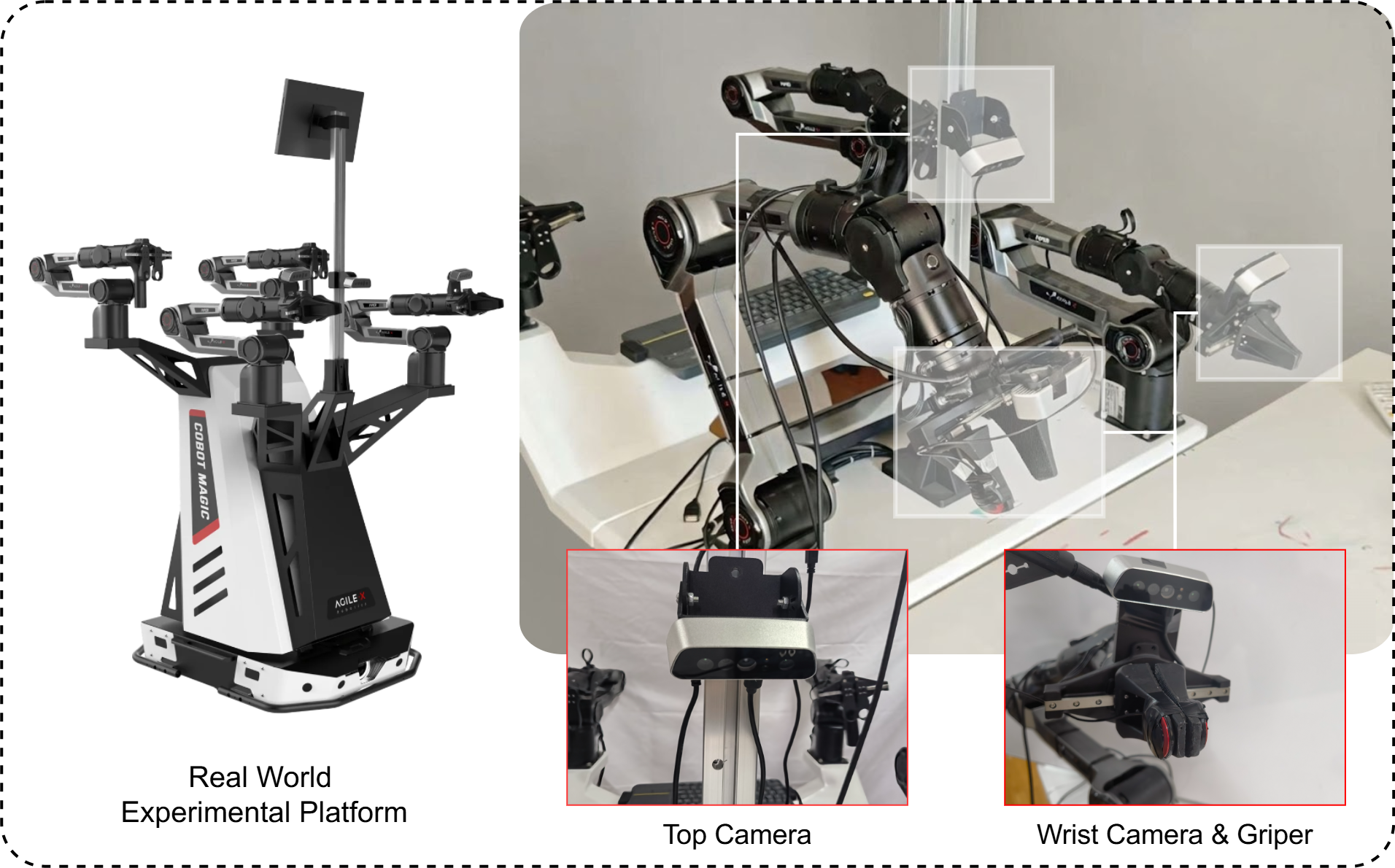}
\caption{Real World Platform.}
\label{fig:real-world}
\end{figure}


\section{Setup Details of Real-World Evaluation}
\begin{figure*}[t]
    \centering
     \includegraphics[width=0.9\textwidth]{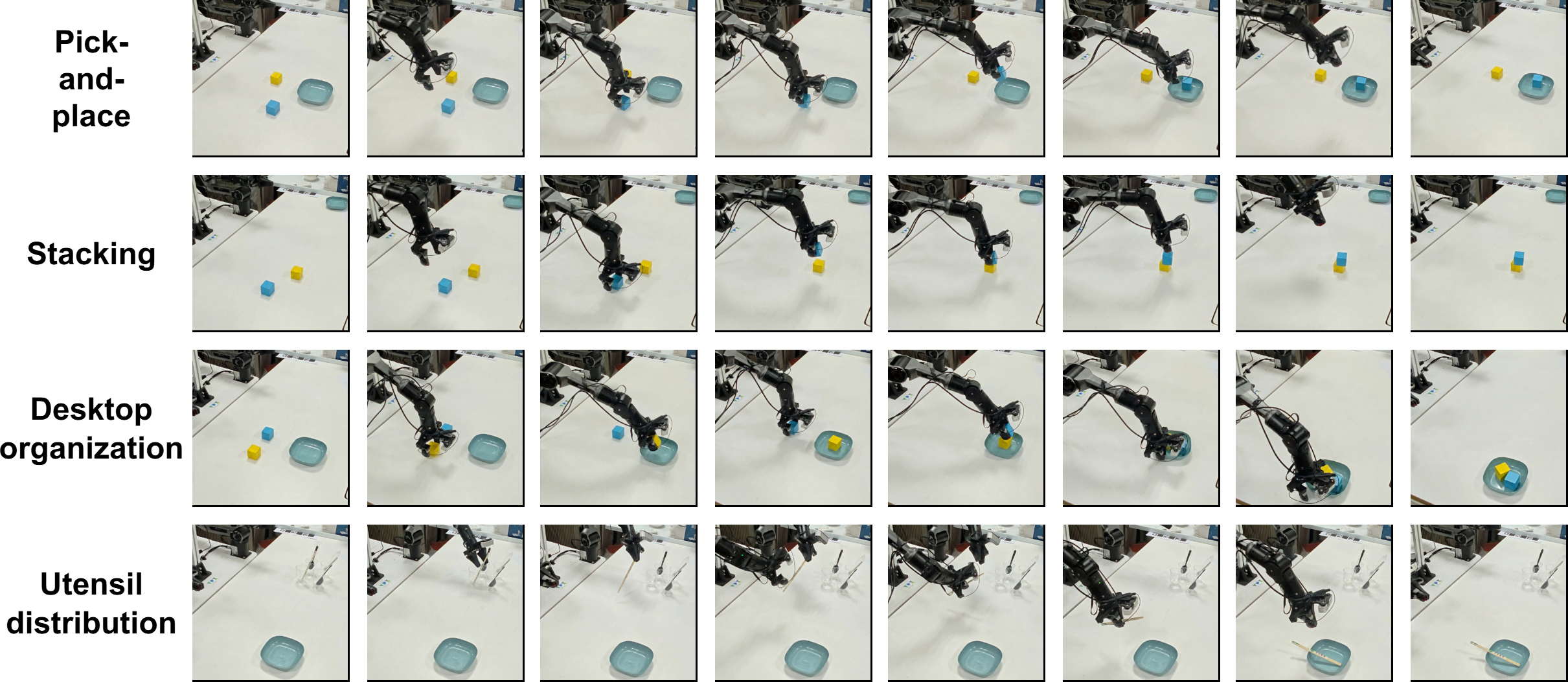} 
    \caption{The four representative tasks in the real-world evaluation.}
    \label{fig:realworldcase}
\end{figure*}

\subsection{Hardware System}
This section provides implementation details of the real-world evaluation, including the hardware platform, the teleoperation interface for expert data collection, task definitions, and the evaluation protocol used to assess S\textsuperscript{2}-VLA. All real-world experiments are conducted on the AgileX Cobot Magic platform. The system is based on the Stanford Mobile ALOHA design~\cite{zhao_learning_2023} and has been integrated and optimized for robust embodied manipulation in real-world environments. The platform consists of three major subsystems, as illustrated in Figure~\ref{fig:real-world}.

\textbf{Bimanual Manipulation Subsystem.} The robot is equipped with two AgileX PiPER 6-DoF robotic arms, each fitted with a parallel two-finger gripper capable of high dynamic response. This design enables stable grasping and precise object manipulation.

\textbf{Mobile Base Subsystem.} The platform integrates an AgileX TRACER differential-drive mobile base, which ensures reliable locomotion and stable operation in unstructured environments.

\textbf{Perception and Onboard Computing.} A multi-view RGB camera system is implemented, including a global-view camera mounted on the head and local-view cameras on both wrists. Onboard inference is performed by an industrial PC equipped with an NVIDIA RTX 4060 GPU and an Intel Core i7 CPU, providing real-time computational support for VLA policy execution.

\subsection{Teleoperation and Data Collection}
We employ a master slave teleoperation interface to collect high-quality expert demonstrations. In this setup, the operator controls the slave robotic arms through a pair of master arms, with accurate motion mapping established between the master devices and the robot end-effectors. The system records multimodal trajectories at 30 Hz, encompassing multi-view RGB observations and arm proprioceptive data.

\subsection{Task Definitions}
We design four representative tasks to comprehensively evaluate the model’s capabilities, as illustrated in Figure~\ref{fig:realworldcase}. For each task, object poses are randomly initialized at the beginning of every trial to test robustness under distributional variations.

\textbf{Pick-and-Place.} The robot identifies and grasps a blue block, then places it into a bowl. This task primarily evaluates basic spatial grounding and generalization ability.

\textbf{Block Stacking.} The robot stacks one block onto another, using the designated blue and yellow blocks. This task demands high precision in end-effector control and stable placement.

\textbf{Desktop Organization.} This is a long-horizon sequential task that involves three steps. First, placing the yellow block into a bowl. Second, placing the blue block into the same bowl. Third, dragging the bowl to the table edge. The task emphasizes multi-stage reasoning and long-horizon execution.

\textbf{Utensil Distribution.} This bimanual coordination task requires the left arm to pick up chopsticks, perform an in-air handover to the right arm, and then the right arm places the chopsticks into a bowl. It assesses fine-grained inter-arm coordination and timing accuracy.\subsection{Implementation and Evaluation Protocol}The implementation details and evaluation protocol for the real-world experiments are summarized as follows.Dataset scale. Demonstration data are collected for each task with the following distribution. 90 demonstrations for the Pick-and-Place task, 90 demonstrations for the Block Stacking task, 150 demonstrations for the Desktop Organization task, and 200 demonstrations for the Utensil Distribution task. \textbf{Preprocessing.} All RGB inputs are resized to a resolution of $224 \times 224$ pixels. The recorded demonstration trajectories are downsampled to 15 Hz for training purposes. \textbf{Inference control rate.} During deployment, the closed-loop inference and control frequency is fixed at 15 Hz. \textbf{Metric.} For each task, 20 independent trials are conducted. The Success Rate (SR) is reported as the primary evaluation metric.

\section{Details of Training and Inference}
\subsection{Inference Procedure}
The inference procedure of S\textsuperscript{2}-VLA operates as a temporally recurrent loop of adaptive fusion and sequential decision-making, as detailed in Algorithm 1. This process requires only 7 GB of VRAM and can be deployed on the NVIDIA RTX 4060 GPU integrated in the ALOHA robotic platform.
\begin{algorithm}[H]
\caption{Inference Process via Adaptive Gating}
\label{alg:inference_clean}
\begin{algorithmic}[1]
\State \textbf{Input:} Initial state $s_0, \mathbf{h}_0^{\text{hid}} = \mathbf{0}$, Horizon $T$
\State \textbf{Output:} Action sequence $\hat{\mathbf{A}}_{t:t+K-1}$
\For{$t=0, \dots, T$} 
    \State \textbf{Step 1: Latent Dynamics}
    \State \quad $(b_t, \mathbf{h}_t^{\text{hid}}) = \text{GRU}(\mathbf{h}_{t-1}^{\text{hid}}, \hat{\mathbf{a}}_{t-1}, P_t)$
    \State \textbf{Step 2: Adaptive Gating}
    \State \quad $\mathbf{g}_t = [g_v, g_c, g_a]^T = \text{Softmax}(\text{MLP}(b_t))$
    \State \textbf{Step 3: Feature Fusion} 
    \State \quad $\mathbf{A}_t^\tau = \sum_{m \in \{v, c, a\}} g_m \cdot \mathbf{O}_m$ 
    \State \textbf{Step 4: Parallel Decoding}
    \State \quad $\hat{\mathbf{A}}_{t:t+K-1} = \text{MLP}_{\text{head}} (\mathbf{A}_t^\tau)$
    
    \State \textbf{Step 5: Execution}
    \State \quad Execute $\hat{\mathbf{a}}_t$, observe $P_{t+1}$
\EndFor
\end{algorithmic}
\end{algorithm}
The key advantage of this inference procedure lies in its state estimation Line 5, which enables the model to maintain temporal consistency and adapt to unexpected perturbations during task execution. The dynamic gating mechanism Line 9 facilitates automatic attention shifting based on the current task phase encoded in the state representation.
\subsection{Training Configurations}

\begin{table}[t]
\centering
\small
\begin{tabular}{ll}
\toprule
\textbf{Setting} & \textbf{Value} \\
\midrule
Optimizer & AdamW \\
Device & NVIDIA H100 GPU $\times$ 4\\
Batch size per device & 16 \\
Max training steps & 100{,}000\\
Warm-up steps & 5{,}000 \\
LR (Action head) & $1\mathrm{e}{-4}$ \\
LR (VLM backbone) & $1\mathrm{e}{-5}$ \\
Scheduler & Cosine annealing \\
Image resolution & $224 \times 224$ \\
\bottomrule
\end{tabular}
\caption{End-to-end training configuration of S\textsuperscript{2}-VLA, including optimizer, learning rate schedules, and computational resources.}
\label{tab:train_settings}
\end{table}
We train S\textsuperscript{2}-VLA end-to-end using the AdamW~\cite{loshchilov_decoupled_2019} optimizer. Since the VLM backbone and the action-generation modules exhibit different sensitivity to optimization dynamics, we adopt a layer-wise learning rate strategy.

\textbf{VLM backbone.} We use a smaller learning rate of $1\mathrm{e}{-5}$ to preserve the stability of pre-trained vision-language representations. \textbf{Action head.} We use a larger learning rate of $1\mathrm{e}{-4}$ to accelerate policy alignment and convergence. We employ a cosine-annealing scheduler with warm-up, where the warm-up stage lasts for the first 5,000 steps. Training is performed on 4 $\times$ NVIDIA H100 GPUs, using a batch size of 16 per device and a total of 100,000 optimization steps. The full configuration is summarized in Table~\ref{tab:train_settings}.
\begin{table}[t]
\centering
\small
\begin{tabular}{ll}
\toprule
\textbf{Hyperparameter} & \textbf{Value} \\
\midrule
Backbone & Qwen3-VL-2B-Instruct \\
Qwen3-VL layer(H) & 28 \\
VLM hidden dim(d) & 2048 \\
Action head hidden dim(d) & 2048 \\
\# SSGAA blocks & 24 \\
\# visual tokens & 128 \\
\# learnable action tokens & 32 \\
Action chunk size(K) & 8 (sim) / 50 (real) \\
action dimension(T) & 7 (single) / 14 (bimanual) \\
\bottomrule
\end{tabular}
\caption{Architecture hyperparameters of S\textsuperscript{2}-VLA.}
\label{tab:arch_hparams}
\end{table}

\subsection{Architecture Hyperparameters}
We use Qwen3-VL-2B-Instruct~\cite{bai_qwen3-vl_2025} as the vision-language backbone, which contains 24 transformer layers (denoted as H). The action head is implemented as a stack of 24 SSGAA blocks. To ensure feature alignment, the hidden dimension of the action head is set to match the backbone hidden size, i.e., 2048 (denoted as d). We use 128 visual tokens and 32 learnable action tokens to encode the multimodal context. For action chunking, we adopt a chunk size of 8 (denoted as K) for LIBERO and SimplerEnv, and 50 for real-world experiments. The generated action dimension is 7-DoF for a single arm (corresponding to position, orientation, and gripper state) and 14-DoF for bimanual control. Key hyperparameters are summarized in Table~\ref{tab:arch_hparams}.

\section{Real-World Gating Analysis}

\begin{figure}[H]
\centering
\includegraphics[width=0.45\textwidth]{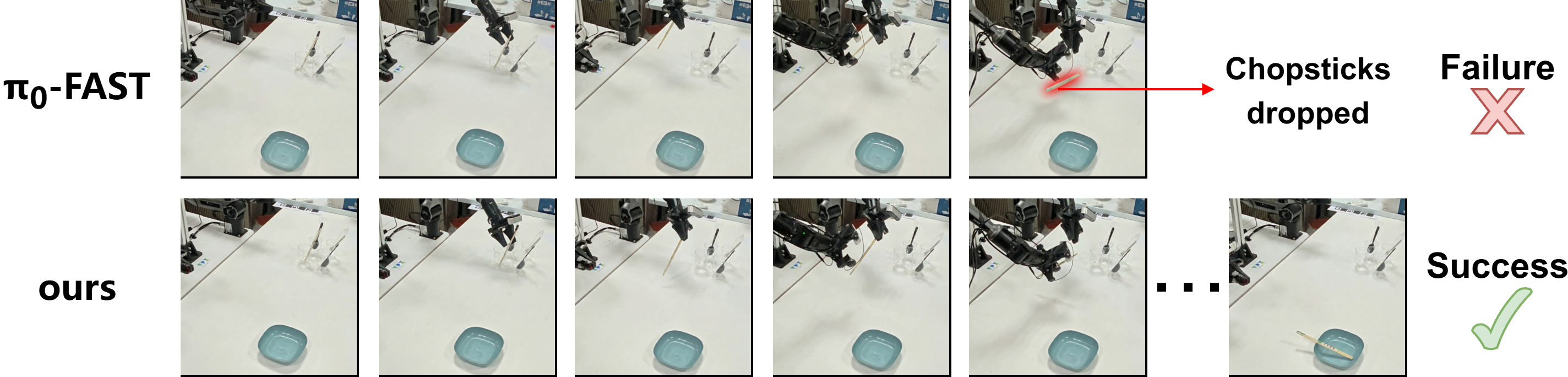}
\caption{S\textsuperscript{2}-VLA exhibits more stable than $\pi_0$-FAST in real tasks.}
\label{fig:real-world-fail}
\end{figure}

To verify the validity of our interpretability findings in real-world scenarios, we execute identical task sequences using both S\textsuperscript{2}-VLA and the $\pi_0$-FAST~\cite{pertsch_fast_2025} model for comparison. As shown in Figure~\ref{fig:real-world-fail}, S\textsuperscript{2}-VLA demonstrates more stable behavior during critical manipulation steps, which aligns with the gating patterns discussed earlier.


\bibliographystyle{named}
\bibliography{ijcai26}